\documentclass{article}
\usepackage{spconf,amsmath,graphicx}
\usepackage{amsfonts}
\usepackage{cleveref}
\usepackage{url}
\usepackage{cite,balance}
\usepackage{textcomp}
\usepackage{amssymb}
\usepackage[ruled,vlined]{algorithm2e}
\usepackage{multirow}
\usepackage{array}

\title{Task-Aware Neural Architecture Search}
\name{Cat P. Le \qquad Mohammadreza Soltani \qquad Robert Ravier \qquad Vahid Tarokh \thanks{This paper was supported by the Office of Naval Research Grant No. N00014-18-1-2244.}}

\address{Department of Electrical and Computer Engineering, Duke University}

\begin{document}
%
\maketitle
\begin{abstract}
The design of handcrafted neural networks requires a lot of time and resources. Recent techniques in Neural Architecture Search (NAS) have proven to be competitive or better than traditional handcrafted design, although they require domain knowledge and have generally used limited search spaces. In this paper, we propose a novel framework for neural architecture search, utilizing a dictionary of models of \emph{base} tasks and the similarity between the \emph{target} task and the atoms of the dictionary; hence, generating an adaptive search space based on the base models of the dictionary. By introducing a gradient-based search algorithm, we can evaluate and discover the best architecture in the search space without fully training the networks. The experimental results show the efficacy of our proposed task-aware approach. 
\end{abstract}

\section{Introduction}
\label{sec:intro}
Neural Architecture Search (NAS) has been a major focal point for work on automated machine learning (AutoML). Initially studied through the lens of reinforcement learning~\cite{zoph2016neural}, a modern development of NAS algorithms largely focuses on minimizing both search time and prior knowledge. Though NAS techniques have greatly improved, many recently proposed methods require significant prior knowledge, e.g. the explicit architecture search domain, or the specific task at hand, as input. This requirement restricts their ability to adapt to situations in which future tasks are potentially unknown.

In this work, we propose a novel, flexible NAS framework, which we call Task-Aware Neural Architecture Search (TA-NAS). The ultimate goal of TA-NAS is to develop an algorithm that dynamically learns an appropriate architecture for a given task at hand, making decisions based on prior history and any information input by the user. Our pipeline is composed of three key components. First, we start with a dictionary of base tasks, the atoms of which consist of architectures that accurately perform said tasks. The dictionary serves as a base on which we dynamically build architectures for new tasks not in the dictionary. Based on the idea that similar tasks should require similar architectures, an often-used assumption in both transfer and lifelong learning, we propose a novel similarity measure for tasks to find the closest base tasks to the new task. Then, we construct a dynamic search space, based on the combined knowledge from the related tasks, without the need for prior domain knowledge. Finally, we present a gradient-based search algorithm, called \emph{Fusion Search} (FUSE). The FUSE algorithm is designed to evaluate the performance of network candidates without fully train any of them. Our experimental evaluation will show the efficacy of our proposed approach.

\section{Related Work}
\label{related}
Many recently proposed NAS techniques have resulted in architectures with performance comparable to those of hand-tuned architectures. The techniques themselves are based on a wide range of techniques, including evolutionary algorithms~\cite{real2018regularized}, reinforcement learning (RL)~\cite{zoph2018learning}, and sequential model-based optimization (SMBO)~\cite{liu2018progressive}. All of these approaches, however, are very time-consuming and need require computational resources, e.g. potentially thousands of GPU-days. To alleviate these issues, differentiable search~\cite{cai2018proxylessnas,liu2018darts,noy2019asap,luo2018neural,xie2018snas} and random search together with sampling sub-networks from a one-shot super-network~\cite{bender2018understanding, li2019random,cho2019one} have been introduced in the literature. For instance, DARTS~\cite{liu2018darts} smooths the architecture search space using a softmax operation. It then solves a bilevel optimization problem which can accelerate the discovery of the final architecture by orders of magnitude~\cite{zoph2016neural, zoph2018learning, real2018regularized, liu2018progressive}. Other recent methods include random search~\cite{li2018massively,li2019random,sciuto2019evaluating,xie2019exploring}, RL based approaches via weight-sharing ~\cite{pham2018efficient}, and network transformations~\cite{cai2018efficient,elsken2018efficient,jaderberg2017population}.

Besides, \cite{bender2018understanding} has thoroughly analyzed the one-shot architecture search using weight-sharing and correlation between the super-graph and sub-networks. None of the above techniques have yet explored the role of the closeness of tasks in the search neural architecture space. Consequently, the search space used by these techniques is often biased and based on the domain knowledge from the well-performed handcrafted neural network architectures. Here, we propose an approach to encode the similarities between tasks for a more efficient search strategy.

\section{Proposed Approach}
\label{sec:NAS}
The pseudocode of TA-NAS is given in Algorithm~\ref{alg1}. At time $t$, we assume that we have access to a dictionary consisting of both previous pairs $(T_{1},X_{1}),...(T_{t-1},X_{t-1}),$ of tasks $T_{k}$ and data sets $X_{k}$ is a given data set, as well as a collection of such pairs that were available upon initialization. Each pair is represented in our dictionary by a trained network. Given the target pair of $(T_{t},X_{t}),$ our goal is to find an architecture for achieving high performance on the target task. In summary, TA-NAS works as follows:
\begin{enumerate}
    \item \textbf{Task Similarity.} Given a new task-data set pair, TA-NAS finds the most related task-data set pairs in the dictionary.
    \item \textbf{Search Space.} TA-NAS defines a suitable search space for the incoming (target) task-data set pair, based on the related pairs.
    \item \textbf{Search Algorithm.} TA-NAS searches to discover an optimal architecture in term of performance for the target task-data set pair on the search space.
\end{enumerate}

\begin{algorithm}[t]
\label{alg1}
\SetAlgoLined
 \textbf{Initialization}: A set of baseline task-data set pairs $B$\;
 \textbf{Input}: Task-data set pairs $(T_{1},X_{1}),...,(T_{K},X_{K})$, Threshold $\tau$, $\epsilon$\;
 \textbf{Output}: Best architecture for the incoming tasks t\;
 
 \For{$t = (K+1), (K+2),...$}{
    \For{$b \in B$}{
        Calculate distance $d_{b,t}$ to find the related tasks\;
        }
    
    Define search space by combining operations, cells, skeleton from related tasks\;
    
    \While{criteria not met}{
        Sample C candidates from search space\;
        Evaluate these candidates using FUSE\;
    }
    
    Add the task t and its architecture to $B.$
 }
 
\caption{Task-Aware NAS}
\end{algorithm}

\subsection{Task Similarity}
\label{ssec:task-taxonomy}
The TA-NAS pipeline heavily depends on the notion of similarity between task-data set pairs. We define the similarity between task-data set pairs in terms of a model-transformation complexity, $N_t$. In particular, we first construct a dictionary with the atoms given by the by trained architectures performing well in each base task-data set pairs. More precisely, let $\ell_{(T,X)}(N)$ be a function that measures the performance of a given architecture $N$ on task $T$ with input data $X$. Here, $\ell_{(T,X)}(N)=1$ denotes the highest performance, while $\ell_{(T,X)}(N)=0$ denotes a complete failure of architecture $N$ on task $T$. For a fixed $0 < \epsilon < 1$, we say that an architecture $N$ is $\epsilon$-representative for $(T,X)$ if $\ell_{T,X}(N) \geq 1 - \epsilon.$ In other words, an architecture $N$ is $\epsilon$-representative if it performs sufficiently well on the given task-data set pair. We thus restrict our attention to $\epsilon$-representative networks. 

We now define our measure of dissimilarity between two task-data set pairs using the notion of $\epsilon$-representative architecture. Let $A =(T_A, X_A)$ and $B=(T_B, X_B)$ be two task-data set pairs, where $N_{A}$ and $N_{B}$ are two trained architectures that are $\epsilon$-representative for $A$ and $B$, respectively. We can define a dissimilarity measure between $A$ and $B$ as follows (hence, the similarity between $A$ and $B$ is given by $1- d^{\epsilon}_{A,B}$):

\begin{figure}[t]
	\centering
	\centerline{\includegraphics[width=8.5cm]{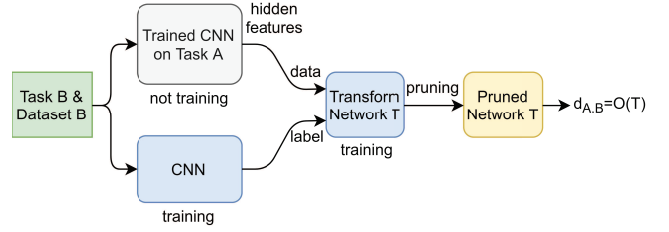}}
	\caption{Illustration of the procedure to compute the distance from task A to task B.}\label{fig: 1}
\end{figure}

\begin{equation}
    d^{\epsilon}_{A,B} = \min_{\{N_{t} \in S_{t} : \ell_{B}(N_t \circ N_A) \geq (1 - \epsilon)\}} O(N_t),
\end{equation}
where $S_{t}$ is a given transform network search space, and $O$ is a general measure of complexity (e.g., the number of parameters in a network). The symbol $\circ$ denotes function (network) composition. In this setting, we assume that the first layer of $N_t$ is the same as the {\emph{penultimate}} layer of $N_A$. In practice, the role of $N_t$ is to transform the features of the penultimate layer of $N_A$ generated by elements of $X_B$ and transform them into elements of $N_B(X_B)$. In other words, the model transformation network $N_t$ should be such that it can transfer the $N_A(X_B)$ into $N_B(X_B)$ with minimum complexity. For instance, if $N_t$ is the identity matrix, it means that $T_A$ and $T_B$ are exactly the same as $N_B(X_B) = N_A(X_B)$. In general, $d^{\epsilon}_{A,B}$ is asymmetric measure, i.e., $d^{\epsilon}_{A,B}\neq d^{\epsilon}_{B,A}$. Ideally, $O$ should be zero for two identical tasks, implying that the architecture for $A$ is also suitable for $B$. In practice, finding the least-complex transform network can be achieved by iterative pruning some super-network for which $N_{t}$ is a sub-graph with the performance as good as the super-network. The dissimilarity measure is the percentage of the non-zero parameters in the pruned $N_t$, and it ranges between 0 and 1. This is illustrated in Figure~\ref{fig: 1}.

\subsection{Search Space}
\label{ssec:search-space}
\begin{figure}[t]
	\centering
	\centerline{\includegraphics[width=6.5cm]{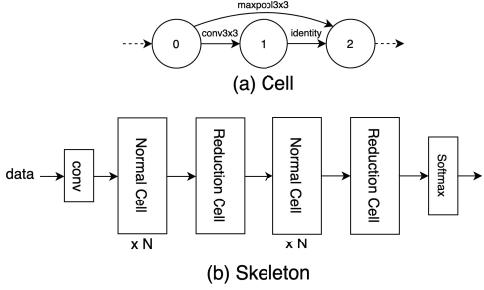}}
	\caption{Illustration of the cell and the skeleton.}\label{fig: 1b}
\end{figure}

Defining a meaningful search space is the key to efficiently finding the best architecture for a specific task. In the NAS literature, the search space is typically defined by stacking a structure called \emph{cell}, as illustrated in Figure~\ref{fig: 1b}. A cell is a densely connected directed-acyclic graph (DAG) of nodes, where all nodes are connected by operations. Other NAS techniques such as one-shot approaches (e.g., DARTS~\cite{liu2018darts}, NAS-Bench201~\cite{dong2020bench}) have also introduced another structure in the search space referred to as \emph{skeleton}. A skeleton is a combination of cells with other operations, forming the complete network architecture. A skeleton is normally predefined, and the goal of NAS algorithms is to find the optimal cells. In this paper, we similarly define the search space in terms of skeletons and cells. Specifically, we focus our search on cells and their operations. As mentioned, cells consist of nodes and operations. Each node has $2$ inputs and $1$ output. The operations (e.g., identity, zero, convolution, pooling) are set so that the dimension of the output is the same as that of the input. If $\mathbf{n}$ is the number of nodes in a cell and $\mathbf{m}$ denotes the number of operations, the total number of possible cells is given by: $\mathbf{m} \times \exp \big({\frac{\mathbf{n}!}{2(\mathbf{n}-2)!}}\big)$.

Our use of a dissimilarity measure gives us knowledge about how related two tasks are. Build upon this knowledge, we can define the search space of the target task-data set pair by combining the skeletons, cells, and operations from only the most similar pairs in the dictionary. Since the search space is restricted to only related tasks, the architecture search algorithm can perform efficiently and requires few GPU hours to find the best candidate network. We have illustrated this in the experimental section.

\subsection{Search Algorithm}
\label{ssec:IDARTS}
The Fusion Search (FUSE) is a novel search algorithm that considers the network candidates as a whole and performs the optimization using gradient descent. Let $C$ be the set of candidate networks on which we define the search space. Given $c \in C$ and training data $X$, denote by $c(X)$ the output of the network candidate $c$. The FUSE algorithm, as illustrated in Algorithm 2, is based on the continuous relaxation of the network outputs. It is capable of searching through all networks in the relaxation space without fully training them. We use as our relaxed space $C$ the set of all convex combinations of candidate networks, which each weight in the combination given by exponential weights:
\begin{equation}
    \Bar{c}(X) = \sum_{c\in C}\frac{\exp{(\alpha_c)}}{\sum_{c'\in C}\exp{(\alpha_{c'})}} c(X),
\end{equation}
where $\Bar{c}$ is the weighted output of network candidate $c$, and $\alpha_c$ is a continuous variable that assigned to candidate $c$'s output. We then conduct our search by jointly training the network candidates and optimizing their $\alpha$ coefficients. Let $X_{train}$, $X_{val}$ be the training and validation data set. The training procedure is based on alternative minimization and can be divided into: (i) freeze $\alpha$ coefficients, jointly train network candidates, (ii) freeze network candidates, update $\alpha$ coefficients. Initially, $\alpha$ coefficients are set to $1/|C|$. While freezing $\alpha$, we update the weights in network candidates by jointly train the relaxed output $\Bar{c}$ with cross-validation loss on training data:
\begin{equation}{\label{eq3}}
    \min_w \mathcal{L}_{train}(w; \alpha, \Bar{c}, X_{train}),
\end{equation}

where $w$ are weights of network candidates in $C$. Next, the weights in those candidates are fixed while we update the $\alpha$ coefficients on validation data:
\begin{equation}{\label{eq4}}
    \min_\alpha \mathcal{L}_{val}(\alpha; w, \Bar{c}, X_{val}).
\end{equation}
These steps are repeated until $\alpha$ converges. The most promising candidate will be selected by: $c^* = \arg\max_{c \in C} \alpha_c$. This training procedure will deliver the best candidate among candidates in $C$ without fully training all of them. In order to go through the entire search space, this process is repeated until certain criteria, such as the number of iterations, the performance of the current most promising candidate, is met.

\begin{algorithm}[t]
\SetAlgoLined
 \textbf{Input}: search space S, $X_{train}$, $X_{val}$, I\;
 \textbf{Initilization}: $c^*$, $\alpha$\;
 \textbf{Output}: Best architecture\;
 
 \For{$i = 1,...,I$}{
    C = [$c^*$, and candidates sampled from S]\;
    \While{$\alpha$ not converge}{
        Update $C$ by descending $\nabla_{w} \mathcal{L}_{train}(w; \alpha, \Bar{c})$\;
        Update $\alpha$ by descending $\nabla_{\alpha} \mathcal{L}_{val}(\alpha; w, \Bar{c})$\;
    }
    $c^* = \arg\max_{c \in C}\alpha_c$\;
 }
 
 \caption{FUSE Algorithm}
\end{algorithm}

\section{Experimental Study}
\label{sec:experiment}
We evaluate the TA-NAS algorithm on image data sets and classification tasks. For our experiment, we initialize the TA-NAS with a set of base binary classification tasks consisting of finding specific digits in  MNIST~\cite{lecun2010mnist} and specific objects in Fashion-MNIST~\cite{xiao2017fashion}. We find $\epsilon$-representatives for each task by pre-training networks on the same architecture ($\text{conv}(32\times5\times5) \rightarrow \text{dense}(1024) \rightarrow \text{dense}(2)$). Here, we pick representative architectures that achieve at least $96\%$ accuracy on their tasks. 

In order to compute the dissimilarity between architectures, we consider for $A$ and $B$ (two task-data set pairs) the first two layers of their trained $\epsilon$-representative networks, which we denote by $N_{A}$ and $N_{B}$, respectively. We then wish to find the least complex architecture that maps hidden features from one task to the other. We thus consider a transform network $N_{t}$ with a  $\text{dense}(2048) \rightarrow \text{dense}(512) \rightarrow \text{dense}(1024)$ architecture. We train $N_{t}$  with mean-square error (MSE) loss on a data set consisting of $N_{A}(X_{B})$ and those of $N_{B}(X_{B})$; here, the goal is to transform $N_{A}(X_{B})$ into $N_{B}(X_{B})$. We then iteratively prune the trained $N_{t}$ as much as possible while maintaining similar performance to $N_{t}$. We take our dissimilarity measure to be the percentage of the remaining non-zero parameters in $N_{t}$ after pruning. We show our results in Figure 2. Our results suggest that two tasks from the same data set (e.g., MNIST or Fashion-MNIST) are often more similar than tasks involving different data sets. It is perhaps interesting to note that the similarity from MNIST tasks to Fashion-MNIST tasks is greater than the similarity from Fashion-MNIST tasks to those in MNIST. Consequently, we can often use Fashion-MNIST knowledge on MNIST, but not vice-versa. 

The task on which we perform NAS is binary classification on Quick, Draw!~\cite{jongejan2016quickdraw} dataset. The Quick, Draw! is a doodle drawing dataset of 345 categories. In this experiment, we select a subset of the Quick, Draw! with a similar format to MNIST and fashion-MNIST, by choosing only 10 categories (e.g., apple, baseball bat, bear, envelope, guitar, lollipop, moon, mouse, mushroom, rabbit) with each has 60,000 data points. Our task of interest is the moon indicator from this subset of Quick, Draw! data set. We pre-train this task and compute dissimilarities to other tasks in the same manner as above. To conduct our search, we choose the top three most similar tasks to the baselines discussed above: (i) digit 0, (ii) trouser, (iii) digit 3 indicators. Due to the similarity in the shape of digit 0 and the moon, the base task of digit 0 indicator is the most related task to the moon indicator.

\begin{figure}[t]
	\centering
	\centerline{\includegraphics[width=6.4cm]{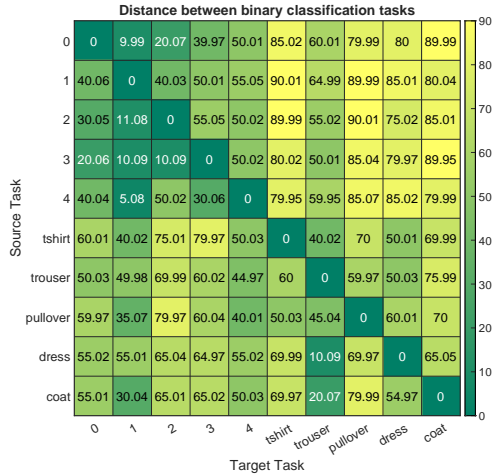}}
	\caption{The distance matrix of baseline tasks.}\label{fig: 2}
\end{figure}

After obtaining the related tasks, we combine the operations and cell structures to generate a suitable search space for the target task. The cell consists of 4 nodes, with 6 edges of operations. The list of operations includes identity, zero, dil-conv3x3, sep-conv3x3, maxpool2x2. Next, the search algorithm is used to find the best architecture in this search space. Initially, three network architectures are randomly generated from the search space. At each iteration, FUSE quickly evaluates these candidates and only saves the best architecture for the next iteration. The search stops only when all criteria (e.g., a prespecified maximum number of iteration or best architecture converges) are met. The results in Table 1 give the best test error of the optimal architecture found by TA-NAS  after 20 trials, in comparison with a random search algorithm and other state-of-art handcrafted networks (e.g., DenseNet~\cite{huang2017densely}, ResNet~\cite{he2016deep}). The architecture produced by TA-NAS is competitive with manual-designed networks while having a smaller number of parameters. When comparing with the random search method using our pre-defined search space, TA-NAS approach achieves a higher accuracy model with less search time in terms of GPU days (i.e., the number of days for a single GPU to perform the task). Hence, this framework can utilize the knowledge of related tasks to find the efficient network architecture for the target task.

\begin{table}
\begin{tabular}{ p{4cm} | p{0.9cm} | p{1.1cm} | p{0.8cm} } 
 \hline
 \multirow{2}{3cm}{\textbf{Architecture}} & \textbf{Error} & \textbf{Param}  & \textbf{GPU}\\
 & \textbf{(\%)} & \textbf{(M)} & \textbf{days}\\
 \hline
 ResNet-18~\cite{he2016deep}                & 1.42 & 11.44 & -\\ 
 ResNet-34~\cite{he2016deep}                & 1.2 & 21.54 & -\\ 
 DenseNet-161~\cite{huang2017densely}       & 1.17 & 27.6 & -\\
 \hline
 Random Search                              & 1.33 & 2.55 & 4\\ 
 FUSE w. standard space                     & 1.21 & 2.89 & 2\\ 
 FUSE w. task-aware space                   & 1.18 & 2.72 & 2\\ 
 \hline
\end{tabular}
\caption{\label{tab:result}Comparison with state-of-art image classifiers on Quick, Draw! dataset.}
\end{table}

\section{Conclusion}
\label{sec:conclusion}
We proposed TA-NAS, a novel task-aware framework to address the Neural Architecture Search problem. By introducing a similarity measure for given pairs of tasks and data sets, we can define a restricted, dynamic architecture search space for a new task-data set pair based on similar previously observed pairs. Additionally, we proposed the gradient-based search algorithm, FUSE, to quickly evaluate the performance of network candidates in the search space. This search algorithm can be applied to find the best way to grow or to compress the current network.


\vfill\pagebreak
\balance
\bibliographystyle{IEEEbib}
\bibliography{refs}

\begin{thebibliography}{10}

\bibitem{zoph2016neural}
Barret Zoph and Quoc~V Le,
\newblock ``Neural architecture search with reinforcement learning,''
\newblock {\em Proc. Int. Conf. Learning Representations}, 2017.

\bibitem{real2018regularized}
Esteban Real, Alok Aggarwal, Yanping Huang, and Quoc~V Le,
\newblock ``Regularized evolution for image classifier architecture search,''
\newblock {\em Proc. Assoc. Adv. Art. Intell. (AAAI)}, 2019.

\bibitem{zoph2018learning}
Barret Zoph, Vijay Vasudevan, Jonathon Shlens, and Quoc~V Le,
\newblock ``Learning transferable architectures for scalable image
  recognition,''
\newblock {\em {IEEE} Conf. Comp. Vision and Pattern Recog}, 2018.

\bibitem{liu2018progressive}
Chenxi Liu, Barret Zoph, Maxim Neumann, Jonathon Shlens, Wei Hua, Li-Jia Li,
  Li~Fei-Fei, Alan Yuille, Jonathan Huang, and Kevin Murphy,
\newblock ``Progressive neural architecture search,''
\newblock {\em Euro. Conf. Comp. Vision}, 2018.

\bibitem{cai2018proxylessnas}
Han Cai, Ligeng Zhu, and Song Han,
\newblock ``Proxylessnas: Direct neural architecture search on target task and
  hardware,''
\newblock {\em Proc. Int. Conf. Learning Representations}, 2019.

\bibitem{liu2018darts}
Hanxiao Liu, Karen Simonyan, and Yiming Yang,
\newblock ``Darts: Differentiable architecture search,''
\newblock {\em Proc. Int. Conf. Machine Learning}, 2018.

\bibitem{noy2019asap}
Asaf Noy, Niv Nayman, Tal Ridnik, Nadav Zamir, Sivan Doveh, Itamar Friedman,
  Raja Giryes, and Lihi Zelnik-Manor,
\newblock ``Asap: Architecture search, anneal and prune,''
\newblock {\em arXiv preprint arXiv:1904.04123}, 2019.

\bibitem{luo2018neural}
Renqian Luo, Fei Tian, Tao Qin, Enhong Chen, and Tie-Yan Liu,
\newblock ``Neural architecture optimization,''
\newblock {\em Adv. Neural Inf. Proc. Sys. (NeurIPS)}, 2018.

\bibitem{xie2018snas}
Sirui Xie, Hehui Zheng, Chunxiao Liu, and Liang Lin,
\newblock ``Snas: stochastic neural architecture search,''
\newblock {\em arXiv preprint arXiv:1812.09926}, 2018.

\bibitem{bender2018understanding}
Gabriel Bender, Pieter-Jan Kindermans, Barret Zoph, Vijay Vasudevan, and Quoc
  Le,
\newblock ``Understanding and simplifying one-shot architecture search,''
\newblock {\em Proc. Int. Conf. Machine Learning}, 2018.

\bibitem{li2019random}
Liam Li and Ameet Talwalkar,
\newblock ``Random search and reproducibility for neural architecture search,''
\newblock {\em arXiv preprint arXiv:1902.07638}, 2019.

\bibitem{cho2019one}
Minsu Cho, Mohammadreza Soltani, and Chinmay Hegde,
\newblock ``One-shot neural architecture search via compressive sensing,''
\newblock {\em arXiv preprint arXiv:1906.02869}, 2019.

\bibitem{li2018massively}
Liam Li, Kevin Jamieson, Afshin Rostamizadeh, Ekaterina Gonina, Moritz Hardt,
  Benjamin Recht, and Ameet Talwalkar,
\newblock ``Massively parallel hyperparameter tuning,''
\newblock {\em arXiv preprint arXiv:1810.05934}, 2018.

\bibitem{sciuto2019evaluating}
Christian Sciuto, Kaicheng Yu, Martin Jaggi, Claudiu Musat, and Mathieu
  Salzmann,
\newblock ``Evaluating the search phase of neural architecture search,''
\newblock {\em arXiv preprint arXiv:1902.08142}, 2019.

\bibitem{xie2019exploring}
Saining Xie, Alexander Kirillov, Ross Girshick, and Kaiming He,
\newblock ``Exploring randomly wired neural networks for image recognition,''
\newblock {\em arXiv preprint arXiv:1904.01569}, 2019.

\bibitem{pham2018efficient}
Hieu Pham, Melody~Y Guan, Barret Zoph, Quoc~V Le, and Jeff Dean,
\newblock ``Efficient neural architecture search via parameter sharing,''
\newblock {\em Proc. Int. Conf. Machine Learning}, 2018.

\bibitem{cai2018efficient}
Han Cai, Tianyao Chen, Weinan Zhang, Yong Yu, and Jun Wang,
\newblock ``Efficient architecture search by network transformation,''
\newblock {\em Proc. Assoc. Adv. Art. Intell. (AAAI)}, 2018.

\bibitem{elsken2018efficient}
Thomas Elsken, Jan~Hendrik Metzen, and Frank Hutter,
\newblock ``Efficient multi-objective neural architecture search via lamarckian
  evolution,''
\newblock {\em Proc. Int. Conf. Learning Representations}, 2019.

\bibitem{jaderberg2017population}
Max Jaderberg, Valentin Dalibard, Simon Osindero, Wojciech~M Czarnecki, Jeff
  Donahue, Ali Razavi, Oriol Vinyals, Tim Green, Iain Dunning, Karen Simonyan,
  et~al.,
\newblock ``Population based training of neural networks,''
\newblock {\em arXiv preprint arXiv:1711.09846}, 2017.

\bibitem{dong2020bench}
Xuanyi Dong and Yi~Yang,
\newblock ``Nas-bench-102: Extending the scope of reproducible neural
  architecture search,''
\newblock {\em arXiv preprint arXiv:2001.00326}, 2020.

\bibitem{lecun2010mnist}
Yann LeCun, Corinna Cortes, and CJ~Burges,
\newblock ``Mnist handwritten digit database,''
\newblock {\em AT\&T Labs [Online]. Available: http://yann. lecun.
  com/exdb/mnist}, vol. 2, pp. 18, 2010.

\bibitem{xiao2017fashion}
Han Xiao, Kashif Rasul, and Roland Vollgraf,
\newblock ``Fashion-mnist: a novel image dataset for benchmarking machine
  learning algorithms,''
\newblock {\em arXiv :1708.07747}, 2017.

\bibitem{jongejan2016quickdraw}
Jongejan J., Rowley H., Kawashima T., Kim J., and Fox-Gieg N.,
\newblock ``The quick, draw! - a.i. experiment,''
\newblock {\em https://quickdraw.withgoogle.com/}, 2016.

\bibitem{huang2017densely}
Gao Huang, Zhuang Liu, Laurens Van Der~Maaten, and Kilian~Q Weinberger,
\newblock ``Densely connected convolutional networks,''
\newblock {\em {IEEE} Conf. Comp. Vision and Pattern Recog}, 2017.

\bibitem{he2016deep}
Kaiming He, Xiangyu Zhang, Shaoqing Ren, and Jian Sun,
\newblock ``Deep residual learning for image recognition,''
\newblock {\em {IEEE} Conf. Comp. Vision and Pattern Recog}, 2016.

\end{thebibliography}

\end{document}